\documentclass{article}
\usepackage{spconf,amsmath,graphicx}
\usepackage{multirow}
\usepackage[colorlinks=true,citecolor=blue]{hyperref}
\title{Learning Instance Representation Banks \\ for Aerial Scene Classification}
\name{Jingjun Yi, Beichen Zhou}%*\thanks{*Corresponding author: q\_bi@whu.edu.cn  \qquad \qquad \qquad \qquad \qquad \qquad   \linebreak This research is supported by the National Key Research and Development Program of China (No.2017YFB0503600) and the National Key Research and Development Program of China (No. 2016YFB0502600).}}
\address{School of Remote Sensing and Information Engineering, Wuhan University, China}
\begin{document}
%\ninept
%
\maketitle
\begin{abstract}
Aerial scenes are more complicated in terms of object distribution and spatial arrangement than natural scenes due to the bird view, and thus remain challenging to learn discriminative scene representation. Recent solutions design \textit{local semantic descriptors} so that region of interests (RoIs) can be properly highlighted. However, each local descriptor has limited description capability and the overall scene representation remains to be refined. In this paper, we solve this problem by designing a novel representation set named \textit{instance representation bank} (IRB), which unifies multiple local descriptors under the multiple instance learning (MIL) formulation. This unified framework is not trivial as all the local semantic descriptors can be aligned to the same scene scheme, enhancing the scene representation capability.
Specifically, our IRB learning framework consists of a backbone, an instance representation bank, a semantic fusion module and a scene scheme alignment loss function. All the components are organized in an end-to-end manner. Extensive experiments on three aerial scene benchmarks demonstrate that our proposed method outperforms the state-of-the-art approaches by a large margin. 
\end{abstract}
\section{Introduction}
\label{sec1}

Earth vision, also known as remote sensing and Earth observation, now becomes a heated topic in the computer vision community \cite{Ding2019Learning,Bi2021LSE}, as nowadays more access is available to acquire an enormous number of aerial images \cite{Tong2019Land}. Aerial image scene classification is a fundamental task for aerial image understanding, and is drawing increasing attention.

Although deep learning has improved the performance of image understanding significantly in the past decade, aerial scene classification is still challenging mainly due to: 1) complicated spatial arrangement from the bird-view \cite{Bi2020A}; 2) more objects irrelevant to the scene scheme are presented from the large-scale imaging sensors \cite{Bi2021LSE,WangQi2018RA}. 

Recently, extensive efforts have been made to improve to tackle this challenge. One of the promising solutions is to highlight the region of interest (RoI) so that more fruitful local semantic information in large-scale aerial images can be learnt \cite{WangQi2018RA,Bi2020RA,Bi2021LSE}. Unfortunately, each local semantic descriptor has its own limitation, and the scene representation capability remains to be boosted. However, how to unify the local semantic information systematically remains unexplored.

In this paper, we push this frontier by designing an \textit{instance representation bank} (IRB) learning framework to unify multiple local semantic descriptors under the classic multiple instance learning (MIL) formulation \cite{Ilse2018Attention,Wang2016Revisiting}. Specifically, the convolutional features are converted into the instance representation before processed by these descriptors. Notably, the combination under the MIL formulation is not trivial, as it allows each descriptor to focus on the same scene scheme so that the key local regions can be highlighted more precisely.

\begin{figure}[t]
    \centering %插入的图片居中表示
	\includegraphics[width=3.4in]{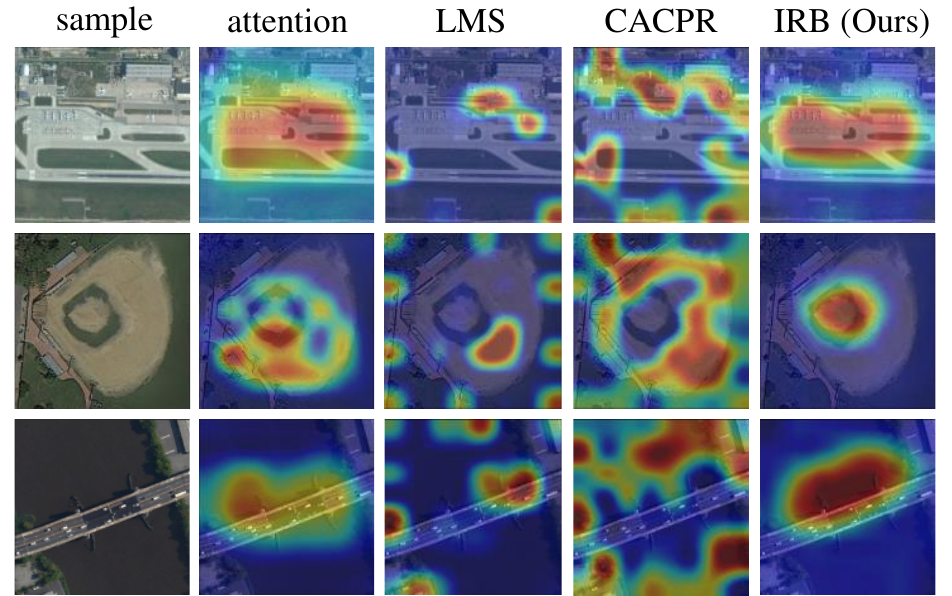}
	%插入的图，包括JPG,PNG,PDF,EPS等，放在源文件目录下
	\caption{Local semantic descriptor based representation visualization for aerial scenes based on attention modules \cite{Bi2020RA,WangQi2018RA,Bi2021MS}, local max selection (LSM) \cite{Bi2019APDC,Bi2021LSE}, context-aware class peak response (CACPR) \cite{Bi2021LSE} and our instance representation bank (IRB) learning. It can be seen that our method can highlight the region of interests more accurately. }  %图片的名称
	\label{fig1}   %标签，用作引用
\end{figure}

Our contribution can be summarized as below.

(1) We propose an instance representation bank (IRB) learning framework for aerial scene classification. Specially, it unifies the latest local semantic descriptors under multiple instance learning formulation while can be embedded into existing deep learning frameworks in a trainable manner. 

(2) We propose a scene scheme alignment loss to align the instance representation banks to the same scene category. Also, a semantic fusion module is proposed to generate scene probability distribution from the instance representation banks. 

(3) Our proposed IRB learning framework achieves the state-of-the-art performance on three widely-utilized aerial scene classification benchmarks.

\section{Related Work}
\label{sec2}
\subsection{Local semantic descriptors for Aerial Scene}
\label{ssec2.1}

Generally speaking, local semantic descriptors can be summarized into three categories, that is, attention based, local max selection based, and class peak response based solutions. 

For attention based local semantic enhancement for aerial scene representation, there were already a flurry of typical works such as \cite{Bi2020RA,WangQi2018RA,Bi2021MS}. However, a major weakness of attention modules in highlighting the key local regions in aerial scenes is its over-activation problem \cite{Bi2020A,ICASSP2021}, bringing more irrelevant features into the final scene prediction. 

On the other hand, local max selection based strategies are also considered and discussed \cite{Bi2019APDC,Bi2021LSE}. The basic idea of such strategies is to select the max feature response in each local window, so that the key regions in aerial scenes can be found. However, as only the local max response is considered, the feature map is usually less interpretable even than the classic convolutional features \cite{Bi2020A,ICASSP2021}. 

More recently, class peak response based local semantic enhancement was used in aerial scene classification. In \cite{Bi2021LSE}, the initial class peak response (CPR) measurement \cite{Zhou2018CPR} in the computer vision community is discussed for its effectiveness in aerial scenes, and a more advanced version \textit{context aware class peak response} (CACPR) is proposed. 

To summarize, each kind of local semantic descriptors has its disadvantage, and how to align them to learn the correct scene scheme with more accurate feature responses still remains an open question. 

\subsection{Multiple Instance Learning}
\label{ssec2.2}

In multiple instance learning (MIL), each object for classification is regarded as a bag, and each bag consists of a series of instances. If a bag contains at least one positive instance, then the bag is judged as a positive bag. Otherwise, the bag is negative \cite{Zhang2004Improve}. Since each instance is only labeled as true or false, MIL is qualified to deal with the weakly-annotated data. Moreover, aggregation functions are needed to transfer instance representations into bag/scene-level representations.

Before the development of deep learning approaches, MIL was usually regarded as a kind of classifier after the feature extraction process \cite{Tang2017Learning,Zhang2016Co}. After the rapid utilization of deep learning approaches, MIL now has the trend to be combined with CNNs in a trainable manner \cite{Wang2016Revisiting,Yang2017MIML}.

To make MIL trainable in deep learning frameworks, recently Ilse et al. assume the bag-level probability distributes as a Bernoulli distribution with the probability $\theta(p)\in[0,1]$ \cite{Ilse2018Attention}. However, the challenge remains as in \cite{Ilse2018Attention} the deep MIL is conducted in the embedding space, where the instance representation fails to generate a bag representation directly. In this case, it needs several fully-connected layers to generate the bag-level (scene-level) probability distribution and the semantic representation capability remains to be enhanced. 

\section{Methodology}
\label{sec3}

\subsection{Framework Overview}
\label{sec3.1}
\begin{figure*}[t]
    \centering %插入的图片居中表示
	\includegraphics[width=6.8in]{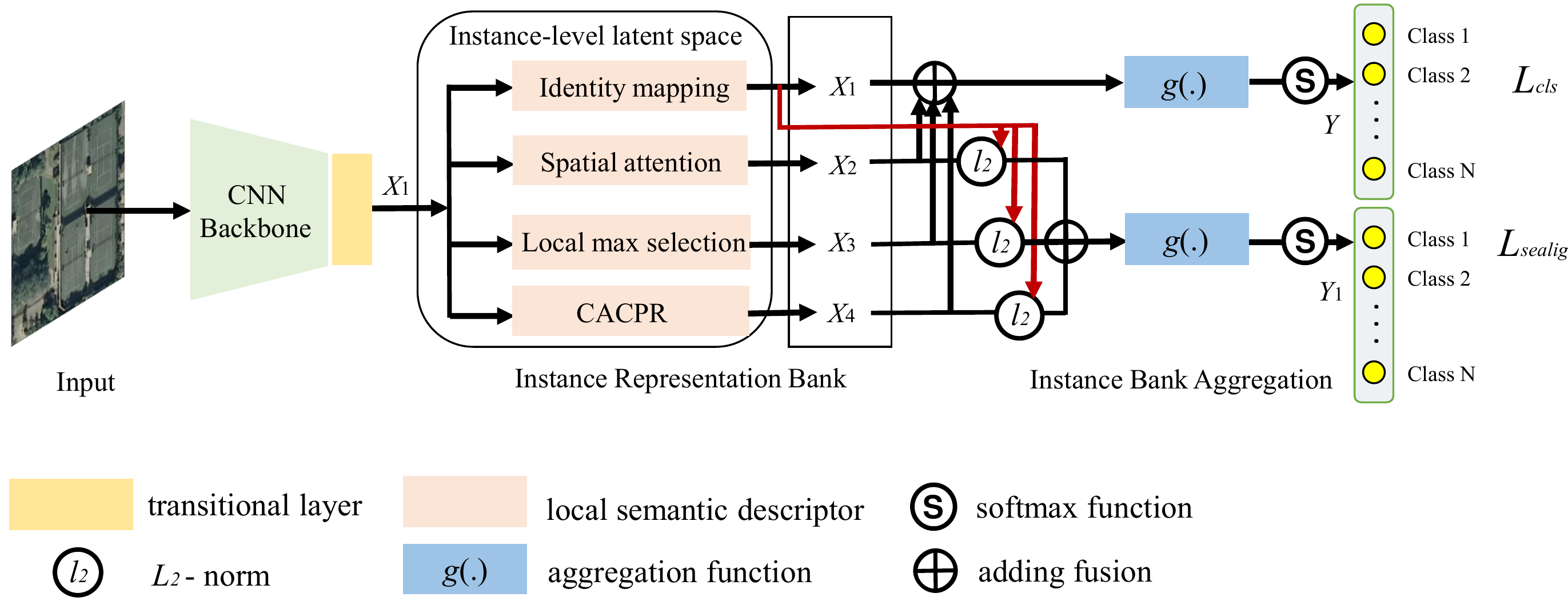}
	%插入的图，包括JPG,PNG,PDF,EPS等，放在源文件目录下
	\caption{Demonstration of our proposed instance representation bank (IRB) learning framework for aerial scene classification.}  %图片的名称
	\label{fig2}   %标签，用作引用
\end{figure*}

Fig.~\ref{fig2}~gives a brief illustration of our proposed IRB learning framework. After converting the traditional convolutional features into an instance representation, multiple local semantic descriptors are utilized to generate the instance representation bank (Sec.~\ref{sec3.2}). Then, a semantic fusion module (Sec.~\ref{sec3.3}) is designed to aggregate each instance representation into a bag probability distribution. Finally, our scene scheme alignment loss (Sec.~\ref{sec3.4}) further allows all the instance representations to align to the same scene scheme.

\subsection{Instance Representation Bank}
\label{sec3.2}
\textit{Instance Representation Transition.} Assume the extracted convolutional features from the ResNet-50 backbone is denoted as $X$, and assume there are overall $N$ scene categories. Then, a convolutional layer with $N$ 1$\times$1 convolutional filters, denoted as $conv1$, is utilized to generate the instance representation $X_{1}$, presented as
$$X_{1}=conv1(X).\eqno{(1)}$$

It is clear that the instance representation $X_{1}$ has $N$ channels so that each channel corresponds to a scene category among the total $N$ categories.

\textit{Local Semantic Descriptor.} Luckily, under the MIL formulation, all three kinds of local semantic descriptors, i.e., attention module, local max selection and class peak response, can be organized as a whole to highlight the key instances in the representation $X_{1}$. 

For the attention module, a naive one-layer spatial attention module with weight $W_{1}$ and bias $b_{1}$ is used to derive the instance representation $X_{2}$ after highlighting, presented as
$$X_{2}=X_{1}(W_{1}X_{1}+b_{1}).\eqno{(2)}$$

For the local max selection operation $L$ \cite{Bi2019APDC}, a sliding window with the size of $W \times W$ is utilized to operate on each channel of the instance representation $X_{1}$. The local semantic enhanced instance representation $X_{3}$ can be calculated via
$$X_{3}=L(X_{1}).\eqno{(3)}$$

For the third kind of local semantic descriptors, we utilize the latest context aware class peak response (CACPR) measure (denoted as CACPR) \cite{Bi2021LSE} to learn the enhanced instance representation $X_{4}$. This process can be presented as 
$$X_{4}=CACPR(X_{1}).\eqno{(4)}$$

Note again: 1) compared with generic CNN frameworks, unifying these latest local semantics descriptors under the instance representation is not trivial, as it helps focus on the same scene scheme; 2) for local max selection and CACPR, we use the default parameter settings in \cite{Bi2019APDC,Bi2021LSE} respectively. 

\textit{Instance Representation Bank.} Under the framework of multiple instance learning, these local semantic descriptors $X_{2}, X_{3}, X_{4}$ and the initial instance representation $X_{1}$ are organized as an instance representation bank (IRB) for learning the semantic probability distribution later, presented as
$$IRB=\{X_{1}, X_{2}, X_{3}, X_{4}\}.\eqno{(5)}$$

\subsection{Semantic Fusion}
\label{sec3.3}
\textit{Instance Bank Aggregation.} All elements inside the instance representation bank (IRB) are summed together to aggregate into the final instance representation $X_{final}$, presented as
$$X_{final}=\sum_{i=1}^{4}X_{i}.\eqno{(6)}$$

\textit{Bag probability distribution generation.} Bear in mind that the final instance representation $X_{final}$ also has $N$ channels and each channel corresponds to a certain scene category. Assume the width and the height of the instance representation is $W$ and $H$ respectively, then the bag-level (scene-level) probability distribution $Y$ is generated as
$$Y=softmax(\sum_{i=0}^{W} \sum_{j=0}^{H} X_{final}), \eqno{(7)}$$
where $softmax$ denotes the softmax normalization. 

\subsection{Scene Scheme Alignment Loss}
\label{sec3.4}
Assume $Y_{i}$ is a certain bag category and there are overall $N$ categories. Then, the classic cross-entropy loss term $L_{cls}$ is presented as
$$L_{cls}=-\frac{1}{N}\sum_{i=1}^N(Y_{i} \log Y_{i}+(1-Y_{i})\log (1-Y_{i})). \eqno{(7)}$$

To eliminate the differences in depicting the scene scheme raised by each element in the IRB, a scene scheme alignment loss is utilized. Firstly, the initial instance representation $X_{1}$ is selected as the base, and the $l$-2 norm $||\cdot||_{2}$ between this base and other elements in the IRB, denoted as $X_{difference}$, is calculated and summed, presented as
$$X_{difference}=\sum_{i=2}^{4} || X_{i}-X_{1} ||_{2}, \eqno{(8)}$$

where \textit{softmax} denotes the softmax classifier. Finally, the cross-entropy loss function is chosen for optimization.
Then it is compressed into the probability distribution $Y_{d}$ by
$$Y_{d}=softmax(\sum_{i=0}^{W} \sum_{j=0}^{H} X_{difference}), \eqno{(9)}$$

\begin{table*}[!t]  
    \centering
    \caption{Comparison of our IRB learning framework and other SOTA approaches (Metrics are presented in \% and are in the form of \textit{'mean accuracy $\pm$ standard deviation'} following the evaluation protocols \cite{Yi2013Geographic,Xia2017AID,Gong2017Remote}.).}
    \begin{tabular}{ccccccc} 
    \hline
    \multirow{2}{*}{Method} & \multicolumn{2}{c}{UCM}&	\multicolumn{2}{c}{AID} &	\multicolumn{2}{c}{NWPU} \\ 
    \cline{2-7}
    ~ & 50\% & 80\% &20\% & 50\% & 10\% & 20\%\\
    %\hline
    %AlexNet \cite{Xia2017AID} & 93.98$\pm$0.67 & 95.02$\pm$0.81 & 86.86$\pm$0.47 & 89.53$\pm$0.31 & 76.69$\pm$0.21 & 79.85$\pm$0.13\\
    %VGGNet-16 \cite{Xia2017AID} & 94.14$\pm$0.69 & 95.21$\pm$1.20 & 86.59$\pm$0.29 & 89.64$\pm$0.36 & 76.47$\pm$0.18 & 79.79$\pm$0.15\\
    %GoogLeNet \cite{Xia2017AID} & 92.70$\pm$0.60 & 94.31$\pm$0.89 & 83.44$\pm$0.40 & 86.39$\pm$0.55 & 76.19$\pm$0.38 & 78.48$\pm$0.26\\
    \hline
    SPP-Net \cite{Han2017Pre} & 94.77$\pm$0.46 & 96.67$\pm$0.94
    & 87.44$\pm$0.45 & 91.45$\pm$0.38 & 82.13$\pm$0.30 & 84.64$\pm$0.23\\
    MIDC-Net \cite{Bi2020A} & 95.41$\pm$0.40 & 97.40$\pm$0.48 & 88.51$\pm$0.41 & 92.95$\pm$0.17 & 86.12$\pm$0.29 & 87.99$\pm$0.18\\
    RA-Net \cite{Bi2020RA} & 94.79$\pm$0.42 & 97.05$\pm$0.48 & 88.12$\pm$0.43 & 92.35$\pm$0.19  & 85.72$\pm$0.25 & 87.63$\pm$0.28\\
    %\hline
    TEX-Net \cite{Rao2017Binary} & 94.22$\pm$0.50 & 95.31$\pm$0.69 & 87.32$\pm$0.37 & 90.00$\pm$0.33 & ---- & ----\\
    %\hline
    D-CNN \cite{Gong2018When} & ---- & 98.93$\pm$0.10& 90.82$\pm$0.16 & \textbf{96.89$\pm$0.10} & 89.22$\pm$0.50 & 91.89$\pm$0.22\\
    %\hline
    MSCP \cite{He2018Remote} & ---- & 98.36$\pm$0.58 & 91.52$\pm$0.21 & 94.42$\pm$0.17 & 85.33$\pm$0.17 & 88.93$\pm$0.14\\
    %\hline
    %FV \cite{Li2017Integrating} & ---- & 98.57$\pm$0.34& ---- & ----& ---- & ----\\
    %\hline
    ARCNet \cite{WangQi2018RA} & 96.81$\pm$0.14 & 99.12$\pm$0.40 & 88.75$\pm$0.40 & 93.10$\pm$0.55 & ---- & ----\\
    DSENet \cite{Wang2021Enhanced} & 96.19$\pm$0.13 & 99.14 $\pm$0.22 & 94.02$\pm$0.21 & 94.50$\pm$0.30 & ---- & ----\\
    DMSMIL \cite{ICASSP2021} & 99.09$\pm$0.36 & 99.45$\pm$0.32 & 93.98$\pm$0.17 & 95.65$\pm$0.22 & 91.93$\pm$0.16 & 93.05$\pm$0.14\\
    %\hline
    \textbf{IRB-Net} (ours) & \textbf{99.14$\pm$0.32} & \textbf{99.67$\pm$0.28} & \textbf{94.25$\pm$0.16} & 96.02$\pm$0.25 & \textbf{92.16$\pm$0.13} & \textbf{93.45$\pm$0.12}\\
    \hline
    \end{tabular} 
     \label{tab1}
\end{table*}

Then, the scene scheme alignment loss term $L_{sealig}$ is presented as
$$L_{sealig}=-\frac{1}{N}\sum_{i=1}^N(Y_{d} \log Y_{d}+(1-Y_{d})\log (1-Y_{d})). \eqno{(10)} $$

The final loss function $L$ is the sum of the above two terms $L_{cls}$ and $L_{sealig}$ respectively, calculated as
$$ L_= L_{cls} + \alpha L_{sealig},  \eqno{(11)}$$

where $\alpha$ is a hyper-parameter to control the impact of the $Y_{d}$. Empirically, $\alpha=5\times10^{-4}$.

\section{Experiment and analysis}
\label{sec4}

\subsection{Benchmark, Evaluation Metrics and Settings}

\textbf{Benchmark.} Three widely-used aerial scene classification benchmarks, namely UC Merced (UCM) \cite{Yi2013Geographic}, Aerial Image Dataset (AID) \cite{Xia2017AID} and Northwestern Polytechnical University (NWPU) \cite{Gong2017Remote}, are utilized to validate our method. 

\textbf{Evaluation Metrics.} Ten independent runs with mean and deviation are required to report for all these benchmarks, and all these benchmarks are divided into training and testing set \cite{Yi2013Geographic,Xia2017AID,Gong2017Remote}. Specially, For UCM, AID and NWPU, the training ratios are 50\% \& 80\% \cite{Yi2013Geographic}, 20\% \& 50\% \cite{Xia2017AID} and 10\% \& 20\% \cite{Gong2017Remote} respectively.

\textbf{Hyper-parameter settings.} ResNet-50 is utilized as our backbone. Adam optimizer is used to train our framework. The pre-trained parameters from ImageNet are used as the initial parameters of our framework to accelerate the training process. Our batch size is set 32, dropout rate is set 0.2, $L_2$ normalization is set $5\times10^{-4}$ and the initial learning rate is $5\times10^{-5}$ while divided by 10 every 20 epochs, until the learning rate reaches $5\times10^{-7}$. For the channel number $N$ in Eq.1, it equals to the category number of each benchmark. 

%\textbf{Development environment.} All our experiments were implemented on a 32GB-memory workstation with Intel(R) Xeon(R) E5-2630 v3 CPU and Titan970Ti GPU. 

\subsection{Comparison with the State-of-the-art Methods}
All the experiments of our IRB and existing state-of-the-art (SOTA) methods on these three datasets under six different experiments are listed in Table.~\ref{tab1}.
It is observed that our method outperforms all the existing SOTA methods. As many of these latest methods focus on local semantic enhancement, the major reason accounting for our superiority may lie in the use of multiple local semantic descriptors while unifying them under the classic multiple instance learning framework. 

\subsection{Ablation Studies}

As our IRB learning framework consists of a ResNet-50 (Res) backbone, an instance representation bank (IRB), a semantic fusion module (SF) and our scene scheme alignment loss (SSA), while in the IRB there are three local semantic descriptors (attention, LMS, CACPR), seven experiments are conducted in our ablation studies on AID benchmark, and the results are listed in Table~\ref{tab2}. It can be observed that all these components contribute positively to scene representation learning. Specifically, the use of multiple local semantic descriptors under MIL framework is more effective than using one of them solely. Also, the scene scheme alignment loss also improves the performance as it aligns instance representations to the same scene scheme.

\begin{table}[!t]  
    \centering
    \caption{Ablation study of our IRB learning framework on AID dataset (Metrics are presented in \% and are in the form of \textit{'mean accuracy $\pm$ standard deviation'} following the evaluation protocols in  \cite{Xia2017AID}.).}
    \begin{tabular}{ccc} 
    \hline
    %\cline{2-3}
    ~ & 20\% & 50\% \\
    \hline
    Res & 88.63$\pm$0.26 & 91.72$\pm$0.17 \\
    \hline
    Res+attention & 89.97$\pm$0.15 & 92.93$\pm$0.15 \\
    Res+LMS & 89.84$\pm$0.19 & 92.57$\pm$0.18 \\
    Res+CACPR & 91.32$\pm$0.17 & 93.54$\pm$0.13 \\
    \hline
    Res+IRB & 92.26$\pm$0.23 & 94.76$\pm$0.17 \\
    %\hline
    Res+IRB+SF & 93.54$\pm$0.18 & 95.53$\pm$0.24 \\
    \hline
    \textbf{Res+IRB+SF+SSA} (ours) & \textbf{94.25$\pm$0.16} & \textbf{96.02$\pm$0.25} \\
    \hline
    \end{tabular} 
     \label{tab2}
\end{table}

\section{Conclusion}
\label{sec5}

In this paper, we propose an instance representation bank (IRB) learning framework for aerial scene classification, which combines the latest local semantic descriptors under the multiple instance learning paradigm. It allows all the descriptors to focus on the same scene scheme, and thus compensate for the weakness of each local semantic descriptor under the existing deep learning framework. Extensive Experiments on three aerial scene classification benchmarks demonstrate the effectiveness of our IRB framework. 

% Below is an example of how to insert images. Delete the ``\vspace'' line,
% uncomment the preceding line ``\centerline...'' and replace ``imageX.ps''
% with a suitable PostScript file name.
% -------------------------------------------------------------------------

\iffalse

\begin{figure}[htb]

\begin{minipage}[b]{1.0\linewidth}
  \centering
  \centerline{\includegraphics[width=8.5cm]{image1}}
%  \vspace{2.0cm}
  \centerline{(a) Result 1}\medskip
\end{minipage}
%
\begin{minipage}[b]{.48\linewidth}
  \centering
  \centerline{\includegraphics[width=4.0cm]{image3}}
%  \vspace{1.5cm}
  \centerline{(b) Results 3}\medskip
\end{minipage}
\hfill
\begin{minipage}[b]{0.48\linewidth}
  \centering
  \centerline{\includegraphics[width=4.0cm]{image4}}
%  \vspace{1.5cm}
  \centerline{(c) Result 4}\medskip
\end{minipage}
%
\caption{Example of placing a figure with experimental results.}
\label{fig:res}
%
\end{figure}
\fi

% To start a new column (but not a new page) and help balance the last-page
% column length use \vfill\pagebreak.
% -------------------------------------------------------------------------
%\vfill
%\pagebreak

\vfill\pagebreak

% References should be produced using the bibtex program from suitable
% BiBTeX files (here: strings, refs, manuals). The IEEEbib.bst bibliography
% style file from IEEE produces unsorted bibliography list.
% -------------------------------------------------------------------------
\bibliographystyle{IEEEbib}
\bibliography{strings,refs}

\end{document}